\documentclass{article}
\usepackage{graphicx}
\usepackage{epstopdf}
\usepackage[numbers]{natbib}
\usepackage{arxiv}
\usepackage[utf8]{inputenc} 
\usepackage[T1]{fontenc}    
\usepackage{hyperref}       
\usepackage{url}            
\usepackage{booktabs}       
\usepackage{amsfonts}       
\usepackage{nicefrac}       
\usepackage{microtype} 
\usepackage{amsmath,amssymb,amsfonts}
\usepackage{algorithmic}
\usepackage{textcomp}
\usepackage{xcolor}

\newtheorem{definition}{Definition}    
    
\newcommand{\bvec}[1]{\mathrm{\mathbf{#1}}}

\title{Learning Humanoid Robot Motions through Deep Neural Networks}

\author{
  Luckeciano C. Melo\\
  Autonomous Computational Systems Lab\\
  Computer Science Division\\
  Aeronautics Institute of Technology \\
  São José dos Campos, Brazil \\
  \texttt{luckeciano@gmail.com} \\
   \And
 Marcos R. O. A. Maximo \\
  Autonomous Computational Systems Lab\\
  Computer Science Division\\
  Aeronautics Institute of Technology \\
  São José dos Campos, Brazil \\
  \texttt{mmaximo@ita.br} \\
  \And
 Adilson Marques da Cunha \\
  Computer Science Division\\
  Aeronautics Institute of Technology \\
  São José dos Campos, Brazil \\
  \texttt{cunha@ita.br} \\
}

\begin{document}
\maketitle

\begin{abstract}
Controlling a high degrees of freedom humanoid robot is acknowledged as one of the hardest problems in Robotics. Due to the lack of mathematical models, an approach frequently employed is to rely on human intuition to design keyframe movements by hand, usually aided by graphical tools. In this paper, we propose a learning framework based on neural networks in order to mimic humanoid robot movements. The developed technique does not make any assumption about the underlying implementation of the movement, therefore both keyframe and model-based motions may be learned. The framework was applied in the RoboCup 3D Soccer Simulation domain and promising results were obtained using the same network architecture for several motions, even when copying motions from another teams.
\end{abstract}


\section{Introduction}

RoboCup Soccer 3D Simulation League (Soccer 3D) is a particularly interesting challenge concerning humanoid robot soccer. It consists of a simulation environment of a soccer match with two teams, each one composed by up to 11 simulated NAO robots \cite{gouaillier2009}, the official robot used for RoboCup Standard Platform League since 2008. Soccer 3D is interesting for robotics research since it involves high level multi-agent cooperative decision making while providing a physically realistic environment which requires control and signal processing techniques for robust low level skills.

In the current level of evolution of Soccer 3D, motion control is a key factor in team's performance. Indeed, controlling a high degrees of freedom humanoid robot is acknowledged as one of the hardest problems in Robotics. Much effort has been devised to humanoid robot walking, where researchers have been very successful in designing control algorithms which reason about reduced order mathematical models based on the Zero Moment Point (ZMP) concept, such as the linear inverted pendulum model \cite{kajita2001}. Nevertheless, these techniques restrict the robot to operate under a small region of its dynamics, where the assumptions of the simplified models are still valid \cite{collins2005,muniz2016}.

Therefore, model-based techniques are hard to use for designing highly dynamic movements, such as a long distance kick and a goalkeeper's dive to defend the goal from a fast moving ball. In the robot soccer domain, a common approach for these movements is to employ keyframe movements, where the motion is composed by a sequence of robot postures. In this case, the movement is designed off-line and executed in a open-loop fashion in execution time.

Due to the lack of mathematical models, an approach frequently employed is to rely on human intuition to design keyframe movements by hand, usually aided by graphical tools. However, this process is difficulty, time consuming, and is often unable to obtain high performance motions given the high dimensionality of the search space. Other possible solution is to use motion capture data from humans \cite{shon2005}, which has its own challenges due to the fact that the kinematic and dynamic properties of a humanoid robot differs greatly from those of a human.

Therefore, many works have experimented on using machine learning and optimization algorithms to develop high performance keyframe movements, showing promising results. Rei et al. describes an algorithm which is able to mimic movements observed from other agents and improve these learnt motions through an evolutionary strategy \cite{depinet2015}. Other works have also used optimization to improve the performance of existing keyframe movements \cite{muniz2016,abbas2017}. Moreover, Peng et al. have developed control policies through deep reinforcement learning that mimic reference motions \cite{peng2018}.

A common fact regarding the aforementioned machine learning and optimization approaches is that they rely on reference motions for learning. Due to the high dimensionality of the search space, direct optimization from a completely random point most likely fails to find an useful movement. In this work, we contribute by showing that a neural network may be taught through supervised learning to mimic an existing movement. From an engineering standpoint, having a keyframe movement represented as a neural network does not provide advantages by itself. However, our intention in a future work is to use this neural network as a seed for reinforcement learning methods as the one shown in \cite{peng2018}.

The remaining of this work is organized as follows. Section \ref{sec:background} provides theoretical background. In Section \ref{sec:methodology}, the methodology and tools used in this work are explained. Furthermore, Section \ref{sec:results_and_discussion} presents simulation results to validate our approach. Finally, Section \ref{sec:conclusion} concludes and shares our ideas for future work.

\section{Background}
\label{sec:background}

\subsection{Keyframe Movements}

\begin{definition}
A \emph{keyframe} \( \mathrm{\mathbf{k}} = \left[ j_1, j_2, \dots, j_n \right]^T \in K \subseteq \mathbb{R}^n \) is an ordered set of joint angular positions, where \( K \) and \( n \) are the joint space and the number of degrees of freedom of the robot, respectively.
\end{definition}

\begin{definition}
A \emph{keyframe step} is an ordered pair \( \mathrm{\mathbf{s}} = \left( \mathrm{\mathbf{k}}, t \right) \in S = K \times \mathbb{R} \), where \( \mathrm{\mathbf{k}} \) is a keyframe and \( t \) is the time when the keyframe must be achieved with respect to the beginning of the movement, respectively. 
\end{definition}

\begin{definition}
A \emph{keyframe movement}, or simply a movement, is defined as \( \mathrm{\mathbf{m}} = \left( \mathrm{\mathbf{s}}_1, \mathrm{\mathbf{s}}_2, \dots, \mathrm{\mathbf{s}}_{\gamma}, r \right) \in M = S^{\gamma} \times \mathbb{R} \), where \( \gamma \) and \( r \) are the number of keyframe steps and the speed rate of the movement, respectively. In this representation, we assume the movement starts at time 0 and the first keyframe step represents the robot posture at the beginning of the movement. Therefore, \( t_1 = 0 \) and each time \( t_i, \forall i \geq 2 \) is a time since the beginning of the movement.
\end{definition}

Keyframe movements are executed in an open-loop fashion, where joint positions are computed through interpolation of keyframe steps based on the current time. If the interface to the robot joints is not position-based, local controllers may be used to track the position references issued by the keyframe. For example, in the Simspark simulator, the simulated NAO has speed-controlled joints, therefore we use simple proportional controllers for each joint to track the desired joint positions. To obtain smooth joint trajectories, we interpolate keyframe steps using cubic splines \cite{bartels1987}, which are functions of class \( \mathcal{C}^2 \). 


\subsection{Neural Networks}
\label{sec:neural_networks}
\begin{figure}[!htbp]
\centering
\includegraphics[width=0.5\textwidth]{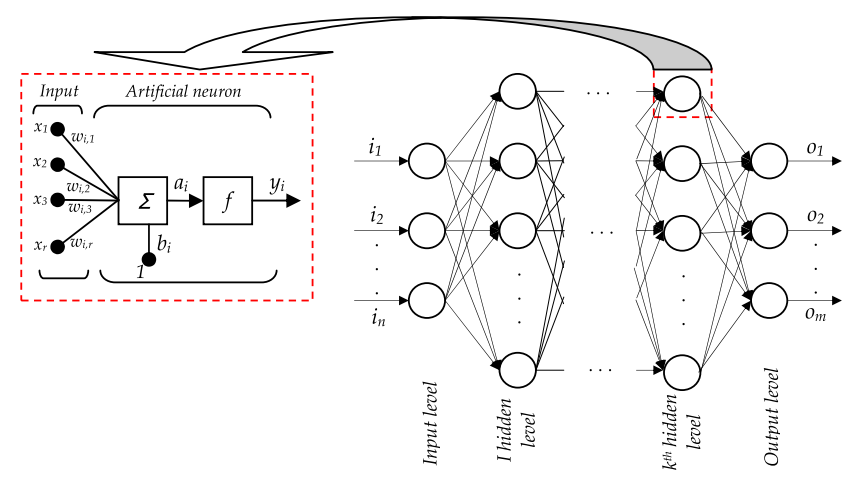}
\caption{Artificial neuron and feed forward artificial neural network \cite{dejan12}. }
\label{fig:ann}
\end{figure}

Neural Networks are a learning representation whose goal is to approximate some function \( f^* \). The data collected from an environment encodes an underlying function \( \mathrm{\mathbf{y}} = f^*(\mathrm{\mathbf{x}}) \) that maps an input \( \textbf{x} \) to an output \( \mathrm{\mathbf{y}} \), which may be a category from a classifier or a continue value in regression problems. The neural network defines an approximate mapping \( \mathrm{\mathbf{y}} = f(\mathrm{\mathbf{x}};\boldsymbol{\theta}) \) by learning the values of the parameters \(\boldsymbol{\theta}\) which result in the best function approximation. The Figure \ref{fig:ann} shows a neural network and an artificial neuron in detail.

These networks are typically represented by composing together many different functions, which are associated with a directed acyclic graph describing a computational model. For example, we might have three layers (each of them representing a function \( f^{(1)}, f^{(2)} \), and \(f^{(3)} \) respectively), connecting in a chain, resulting in a final representation \( f(\mathrm{\mathbf{x}}) = f^{(3)}(  f^{(2)} ( f^{(1)}(\mathrm{\mathbf{x}}))) \).

During neural network training, the objective is to adjust \(f(\mathrm{\mathbf{x}})\) to match \(f^{*}(\mathrm{\mathbf{x}})\) using the training dataset, which provides noisy examples of \(f^{*}(\mathrm{\mathbf{x}})\) evaluated in different points. The training examples specify directly what the output layer must do at each point \(\mathrm{\mathbf{x}}\), but the learning algorithm must decide how to use all layers to produce this desired output \cite{Goodfellow-et-al-2016}.

Additionally, we must also choose a learning algorithm to tune this function approximation. In the context of neural networks, gradient-based algorithms are broadly used, especially those based on the backpropagation idea \cite{hinton88}. The purpose of these algorithms are to propagate the gradient of a cost function through the whole network, in order to minimize the cost function. Most modern neural networks perform this optimization strategy using maximum likelihood, i.e. the cross-entropy between the training data and the model distribution:

\begin{equation}
J(\boldsymbol{\theta}) = -\mathbb{E}_{\mathrm{\mathbf{x}},\mathrm{\mathbf{y}}\sim \hat{p}_{data}}\log{p_{model}(\mathrm{\mathbf{y}} | \mathrm{\mathbf{x}})}
\label{eq:cost_function_ml}
\end{equation}

In this work, we used the mean squared error loss function in order to fit the dataset. Indeed, we may show  that both cost functions are closely related. Let us consider normally distributed errors:

\begin{equation}
{p_{model}(\bvec{y} | \bvec{x})} = \mathcal{N}( \bvec{y}; f (\bvec{x}; \boldsymbol{\theta}), \sigma^{2}\bvec{I})
\label{eq:errors}
\end{equation}
where \(f (\bvec{x}; \boldsymbol{\theta})\) and \(\sigma^{2}\bvec{I}\)  are the mean and covariance of this distribution, respectively. Substituting Eq. \eqref{eq:errors} in Eq. \eqref{eq:cost_function_ml}:

\begin{equation}
J(\boldsymbol{\theta}) = \frac{1}{2}\mathbb{E}_{\bvec{x},\bvec{y}\sim \hat{p}_{data}} \lVert \bvec{y} - f (\bvec{x}; \boldsymbol{\theta}) \rVert ^{2} + const
\label{eq:cost_function_expectation}
\end{equation} 

The constant term does not depend on \( \boldsymbol{\theta} \) and may be dropped. By explicitly evaluating the expectation in Eq. \eqref{eq:cost_function_expectation}, we arrive at the mean squared error cost function:

\begin{equation}
J(\boldsymbol{\theta}) = \frac{1}{2m} \sum^{m}_{i} \lVert y_{i} - f (\bvec{x}; \boldsymbol{\theta}) \rVert ^{2}
\end{equation}

Lastly, the gradient of the loss function is taken and propagated through the hidden layers by the chain rule. For example, given \(\textbf{Y} = g(\textbf{X}) \) and \(z = f(\textbf{Y}) \), then the chain rule states:

\begin{equation}
\nabla_{\textbf{X}}z = \sum_{j}(\nabla_\textbf{X}Y_{j})  \frac{\partial{z}}{\partial{Y_{j}}}
\end{equation}

This equation is taken recursively until the gradient is propagated  to all layers of the neural network.

\section{Methodology and Tools}
\label{sec:methodology}

\subsection{Dataset}\label{AA}
In order to use supervised learning for learning keyframe motions using neural networks, we first need to construct a dataset. A dataset consists of samples of keyframe steps. The samples were collected within Soccer 3D environment with a frequency of 50 Hz. We acquired these samples in two different ways.

In the first one, we commanded an agent of our team to execute specific motions and sampled the reference joint positions computed by our code. In this case, we sampled the kick and get up keyframe motions \cite{muniz2016}. Notice that for this approach to be successful, one needs access to the source code.

The second approach involved changing the Soccer 3D server source code to provide current joint positions of a given robot, in a similar way as described in \cite{macalpine2013}. This allowed us to acquire motion datasets from other teams, without any knowledge of how these movements are implemented. In this case, we collected two types of kicks based on keyframes and sampled joint values of the walking engine \cite{AAAI12-MacAlpine}.

\subsection{Neural Network Architecture and Hyperparameters}

\begin{figure}[!htbp]
\centering
\includegraphics[width=0.5\textwidth]{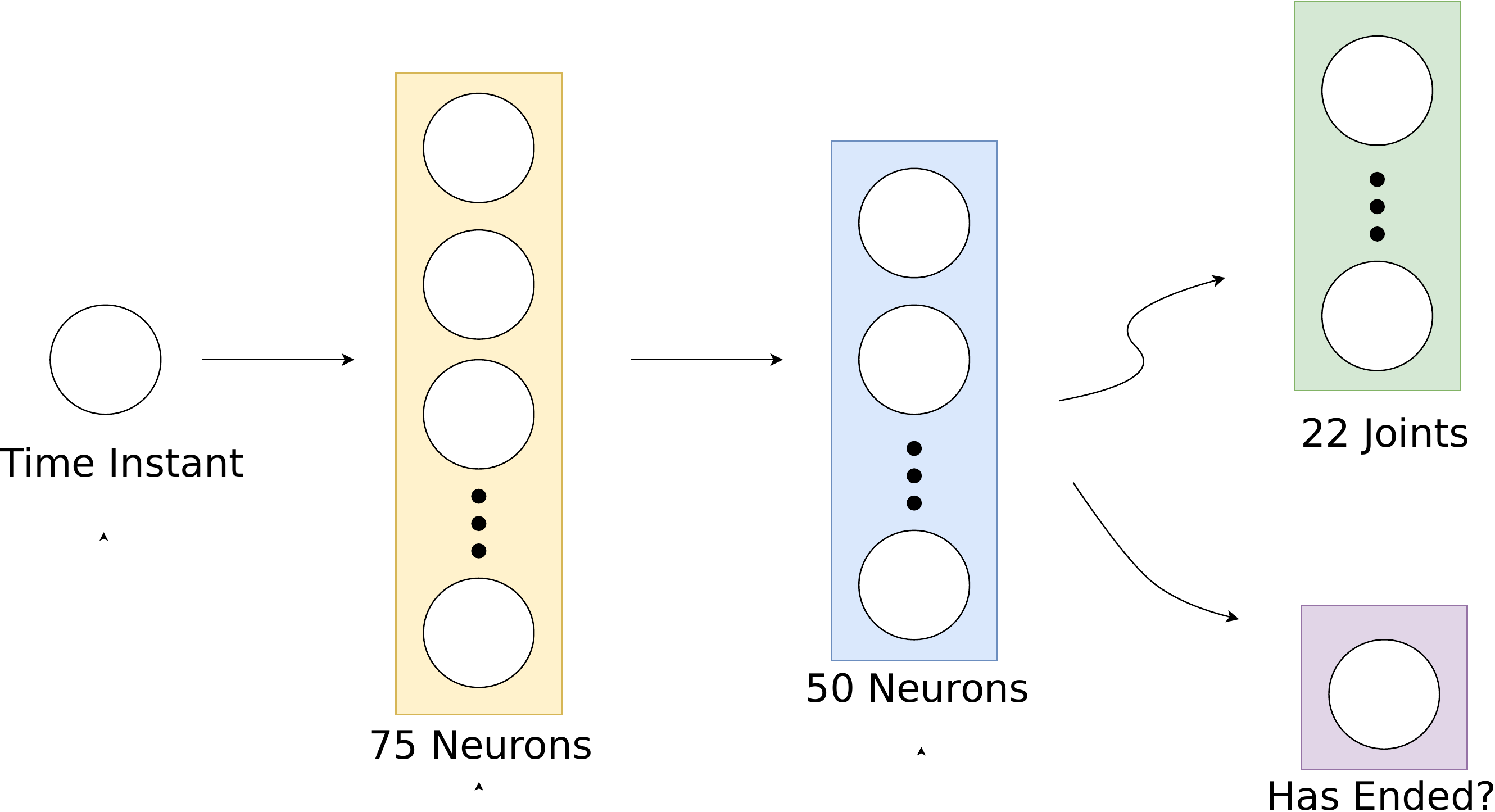}
\caption{Architecture of the neural network designed to learn motions.}
\label{fig:model_plot}
\end{figure}

The neural network has to be able to learn how to interpolate between samples, which actually happens. The architecture that performed best -- in terms of mean absolute error minimization and simplicity -- is shown in Figure ~\ref{fig:model_plot}. A deep neural network with 2 hidden, fully connected layers of 75 and 50 neurons was used. The output layer has 23 regression neurons, which represent the 22 joint angles and a neuron whose output indicates if the motion has ended or not. The neurons in each hidden layer use the LeakyReLU activation function \cite{leakyrelu}: 

\[
  f(x) = \left\{
     \begin{array}{@{}l@{\thinspace}l}
       \alpha x,   & \quad x < 0  \\
       x, & \quad x \ge0 \\
     \end{array}
   \right.
\]
where $\alpha$ is a small constant. This activation function was used to improve the representation capacity of the neural network, adding support for non-linear functions.

This architecture resulted in thousand of parameters to optimize, as exposed in Table \ref{tab:network_summary}. A very high number when compared to more traditional optimization approaches \cite{AAAI12-MacAlpine}. Notice that increasing the number of parameters usually allows representing better movements.

\begin{table}[htbp]
\caption{Network Summary}
\begin{center}
\begin{tabular}{|c|c|c|c|}
\hline
\textbf{Layer}&{\textbf{Neurons}}& \textbf{Activation}& \textbf{Parameters} \\
\hline
Dense & 75 & LeakyReLU & 150  \\
\hline
Dense & 50 & LeakyReLU & 3800 \\
\hline
Dense & 23 & Linear & 1173 \\
\hline
\end{tabular}
\begin{tabular}{|c|c|}
\hline
\textbf{Total Parameters} & 5123 \\
\hline
\end{tabular}
\label{tab:network_summary}
\end{center}
\end{table}

\subsection{Training Procedure}
Since keyframe motions are executed in an open-loop fashion, the sequence of joint positions are always the same for different repetitions, independently of the robot's state. Therefore, adding samples of multiple executions of the same motion would not make our dataset richer, so we decided to use only one repetition for each movement for faster training. In the case of the walking motion, we collected samples within one walking period.

During training, we used 50 thousands epochs divided in 5 training phases, where the learning rate was decreased between phases in order to achieve better performance. First, we executed 30000 epochs using learning rate of 0.001. The other phases had 5000 epochs each, and we decreased the learning rate by 0.0002 in each phase.

Furthermore, we used Adam optimization \cite{adam2014} during the whole training. The loss function used was the mean squared error, as explained in Subsec. \ref{sec:neural_networks}. We decided this loss function is adequate for this problem because it strongly penalizes large errors, which can collapse the whole motion.

\subsection{Deployment in Soccer 3D Environment}
In order to perform network design and the training procedure, we used the Keras \cite{chollet2015keras} framework coupled with Tensorflow \cite{tensorflow2015-whitepaper} as backend. After training, the weights were freezed and converted to a specific format which is readable using the Tensorflow C++ API integrated within the agent's code. Hence, the training is performed outside the environment, but the agent actually computes network inferences during simulation execution.

\section{Results and Discussion}

\label{sec:results_and_discussion}
\subsection{Training Results}

All results and logs obtained, as well as the code used for neural network training, are available in the project repository\footnote{\label{repository_walk} Repository: https://goo.gl/MjRWAH} for reproducibility.

\begin{figure}[!htbp]
\centering
\includegraphics[width=0.6\textwidth]{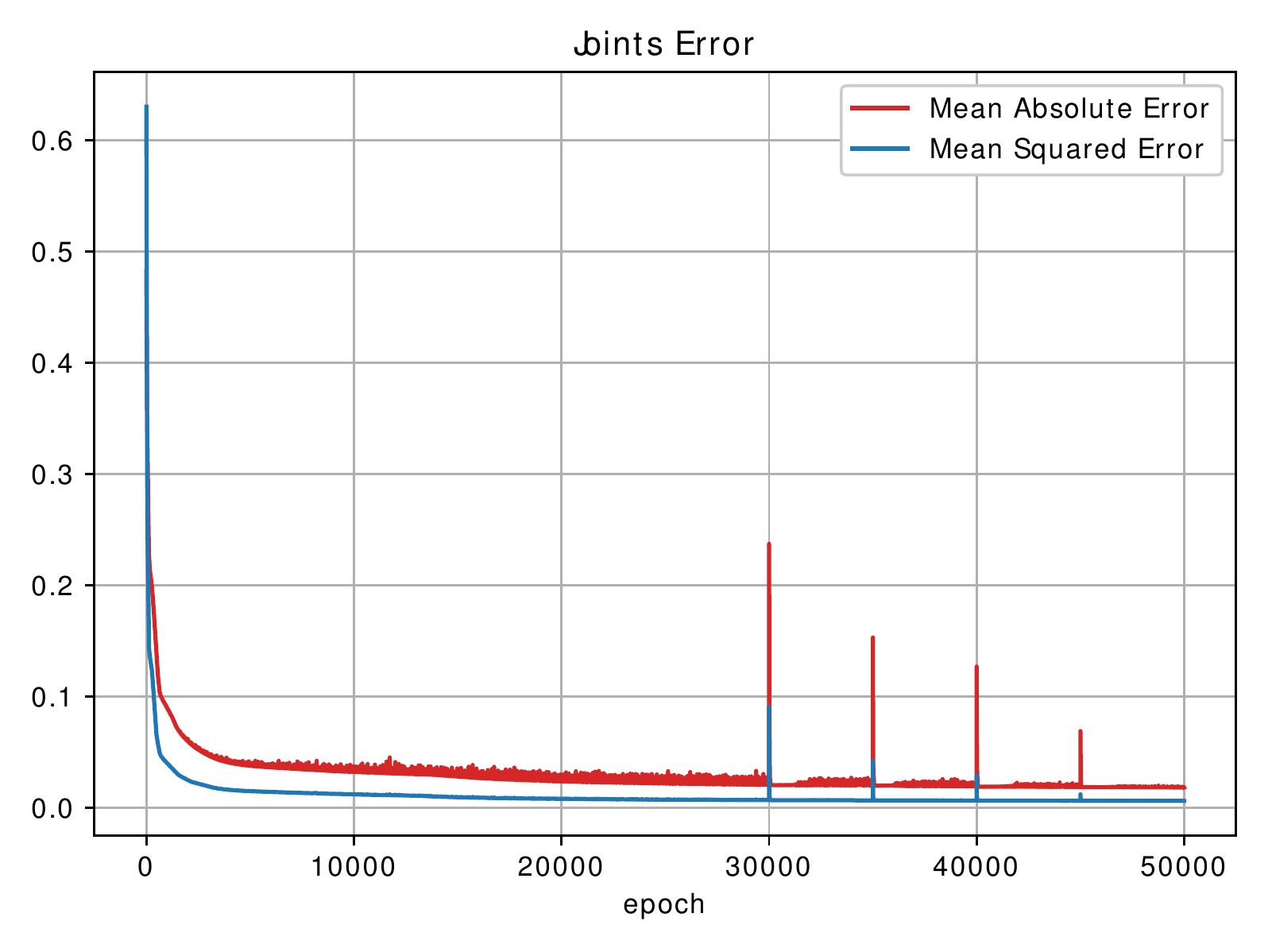}
\caption{The plots of mean squared error and mean absolute error during training.}
\label{fig:errors}
\end{figure}

The initial results come from the training procedure, outside the simulation environment. Figure \ref{fig:errors} presents training curves for the kick keyframe dataset. In this case, the plots show mean squared error and mean absolute error metrics, respectively. In both metrics, the value decreases drastically in the first epochs. This same behavior was present in other training procedures as well. However, only after thousands of epochs the network achieved a low error that reproduced the motion successfully, which shows how sensible to small joint errors keyframes are, given that they are open-loop motions. The peaks during the training happen at the learning rate transition instants, but they do not hurt the training procedure itself. This is due to the fact that we re-compile the model at each training phase, which resets the optimizer state. This means that the training will suffer a little at the beginning until you adjust the learning rate, the moments, etc. However, there is absolutely no damage to the weights.

\subsection{The Learned Kick Motion}

\begin{figure}[!htbp]
\centering
\includegraphics[angle=90,width=0.7\textwidth]{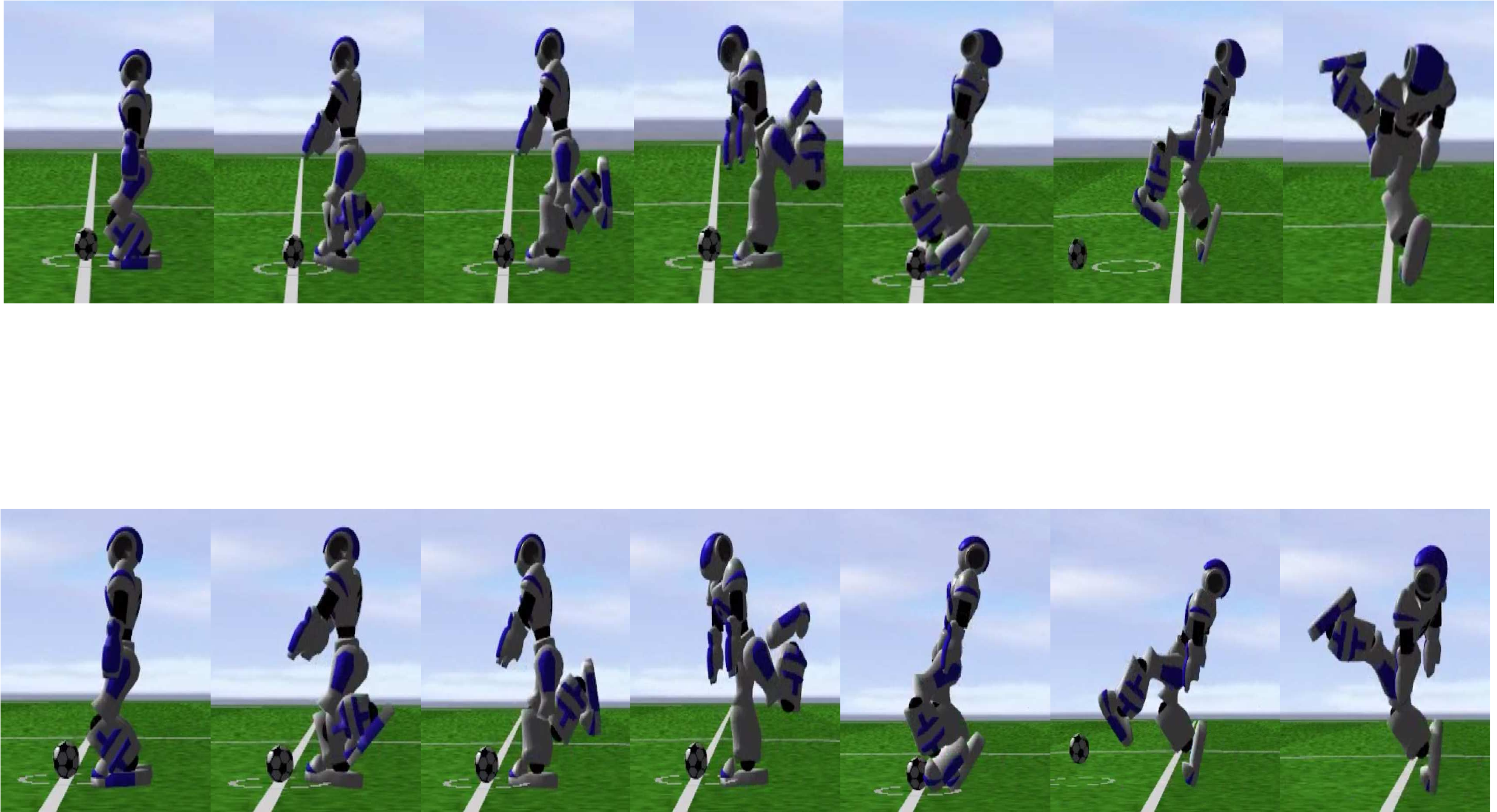}
\caption{Kick motion. The first row of figures shows the original kick motion. The second row shows the learned kick motion. The motions are visually indistinguishable.}
\label{fig:motions}
\end{figure}

\begin{figure*}[!htbp]
\centering
\includegraphics[width=0.7\textwidth]{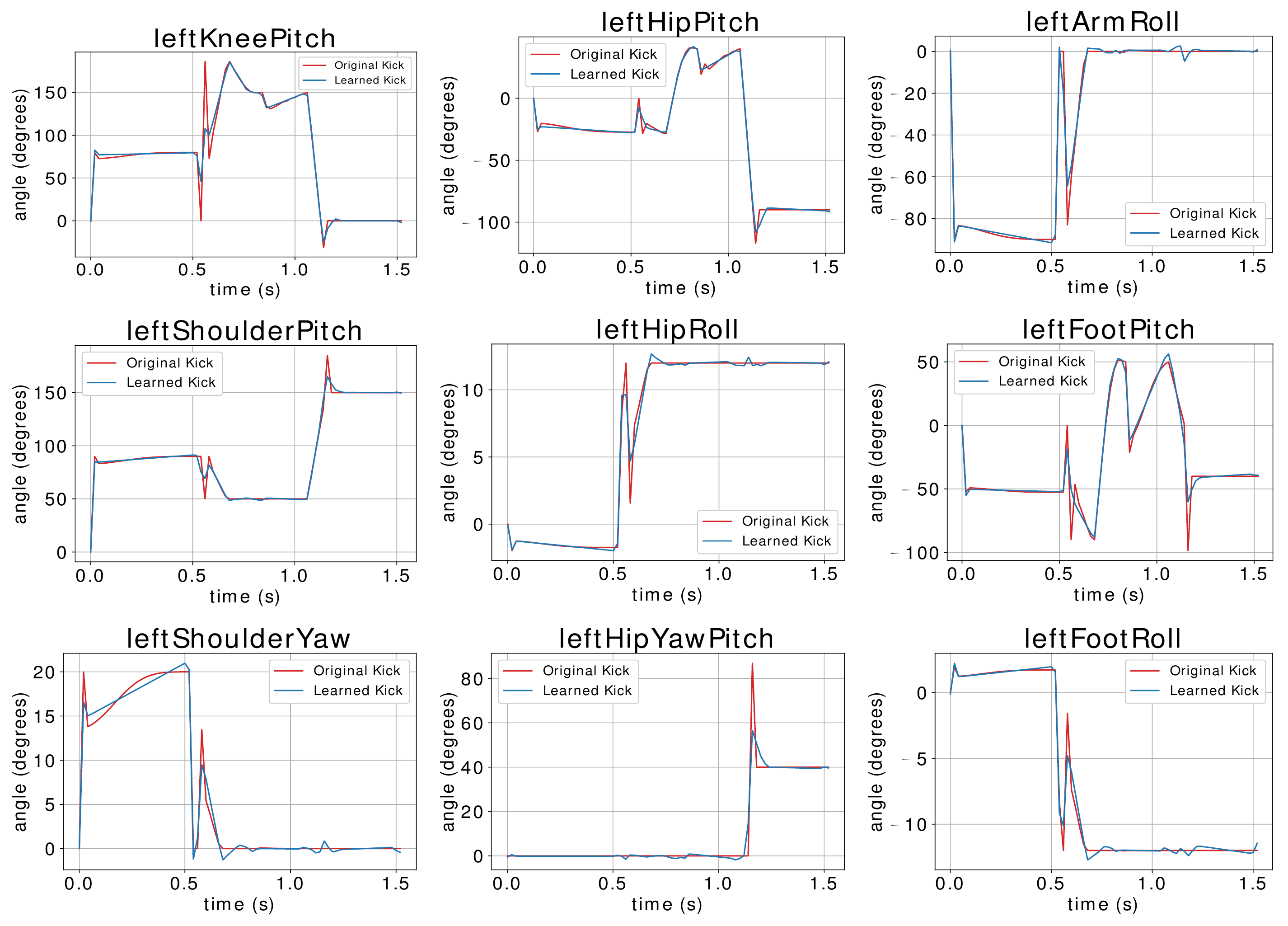}
\caption{Joint values for comparing original and learned kicks. The neural network was able to fit the joint trajectories with small errors.}
\label{fig:kick_joints_curves}
\end{figure*}

The final mean absolute error is \textbf{0.018} radians and the motion is visually indistinguishable from the original one, as can be seen in Figure \ref{fig:motions}. In this figure, snapshots from both motions were taken. The Figure \ref{fig:kick_joints_curves} shows several plots of joint angles comparing the original and learned kick motions. As we may see, the learned motion has fitted the movement with minor errors\footnote{\label{footnote_kick} Kick results video: https://streamable.com/gpltm}.

In order to evaluate the learned kick motion in the RoboCup Soccer 3D domain, we created a statistical test. Inside the test scenario, the ball was placed initially in the center of the field with an agent near to it. The only action of the agent is to kick the ball in the goal direction. After the kick, the agent runs until reaching the ball and kicks it again, repeating this process till scoring a goal. When the goal occurs, this same scenario is repeated. The whole test was conducted during thirty minutes in clock time and the following data was collected: total number of kicks, number of successful kicks, mean distance that the ball has traveled and the standard deviation of this measure. The results from the original and learned kicks is shown in Table \ref{tab_kicks_statistics}.

\begin{table}[htbp]
\caption{Kick Comparison}
\begin{center}
\begin{tabular}{|c|c|c|c|}
\hline
\textbf{Kick}&\multicolumn{3}{|c|}{\textbf{Statistics}} \\
\cline{2-4} 
\textbf{Type} & \textbf{\textit{Accuracy (\%)}}& \multicolumn{2}{|c|}{\textbf{Distance (\(m\))}} \\ 
\cline {3-4}
& & \textbf{\textit{Mean}}& \textbf{\textit{Std}} \\
\hline
Original Kick & 64.5 & 8.92 & 3.82  \\
\hline
Neural Kick & 52.6 & 7.16 & 4.06 \\
\hline
\end{tabular}
\label{tab_kicks_statistics}
\end{center}
\end{table}

Although both kicks have similar results, the original kick is slightly better in this scenario. Confronting Figure \ref{fig:kick_joints_curves}, we can conclude that even with an almost equal representation, the kick lose part of its efficiency and this fact show us how sensible are movements based on keyframe data.

\subsection{The Learned Walk Motion}
Using the modified server described in Subsec. \ref{AA}, a dataset with samples of the UT Austin Villa's walking motion \cite{macalpine2013} was acquired. This team is the current champion of RoboCup Soccer 3D competition \cite{macalpine2017}.

The objective is to mimic the walk motion as a keyframe and use that in our agent. The previously described framework used for learning our own kick motion was used in this training, including the neural network architecture and its hyperparameters.

\begin{figure*}[!htbp]
\centering
\includegraphics[width=0.7\textwidth]{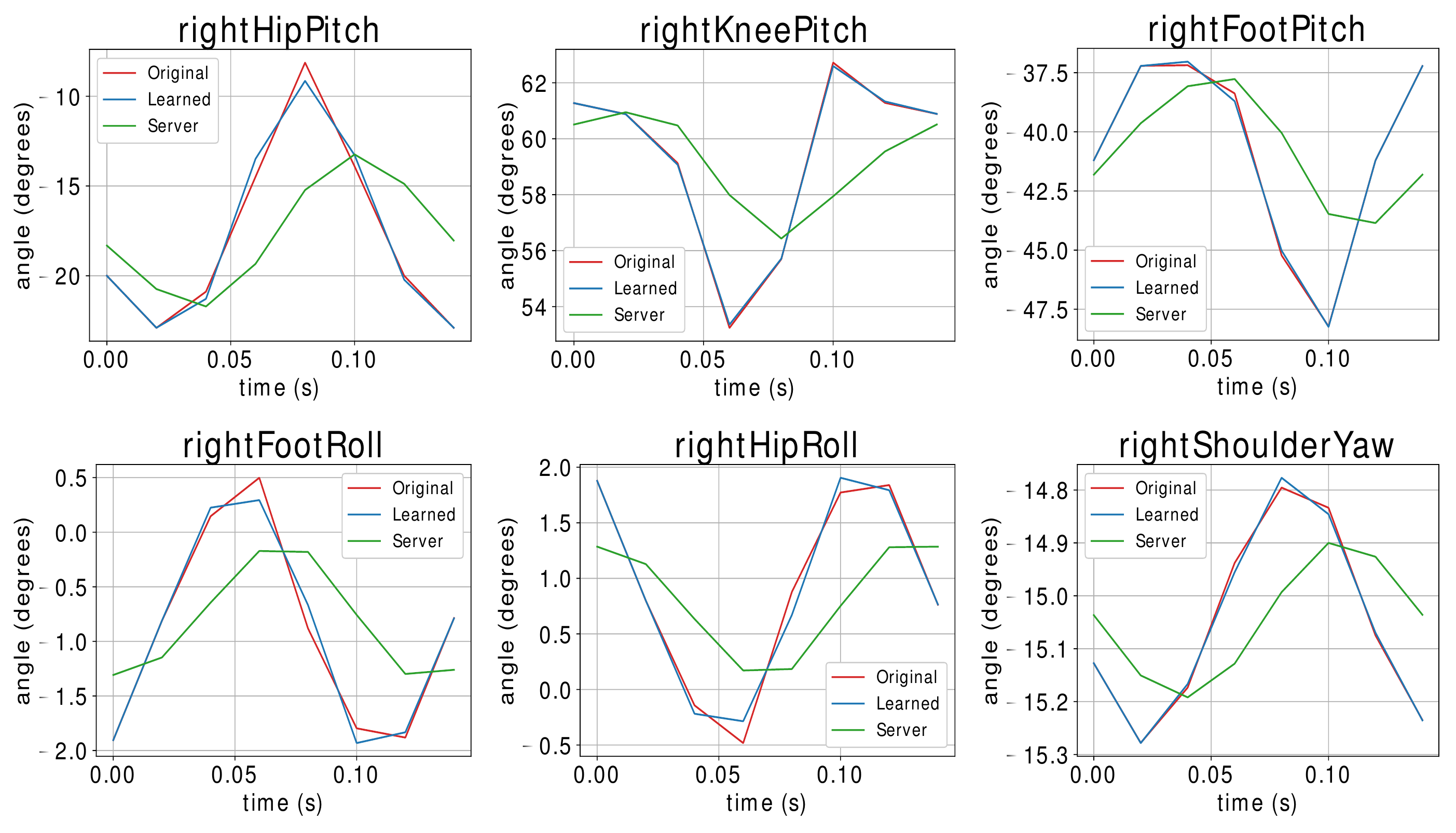}
\caption{Joints positions during a period of the walking motion for the original walk, the learned walk and the joints positions effectively attained during the learned walking motion.}
\label{fig:walk_joints_curves}
\end{figure*}

The results from this training are shown in Figure \ref{fig:walk_joints_curves}. Similarly to Figure \ref{fig:kick_joints_curves}, it shows the joint angles throughout the walking motion period for the original and learned walk. Additionally, it shows the real joints values from the movement in the server. These joints were chosen because they are the most dynamic in the walk motion and therefore the hardest to learn.

The learned motion has fitted the dataset very well. However, these values are just desired joints. In fact, these values are used as references to joint controllers and are attenuated due to joint dynamics. Furthermore, this motion is operated in a open-loop fashion, so the agent is not able to correct its own trajectory, and this walks gets biased in the simple task of walking straight forward.

\begin{figure}[!htbp]
\centering
\includegraphics[width=0.4\textwidth]{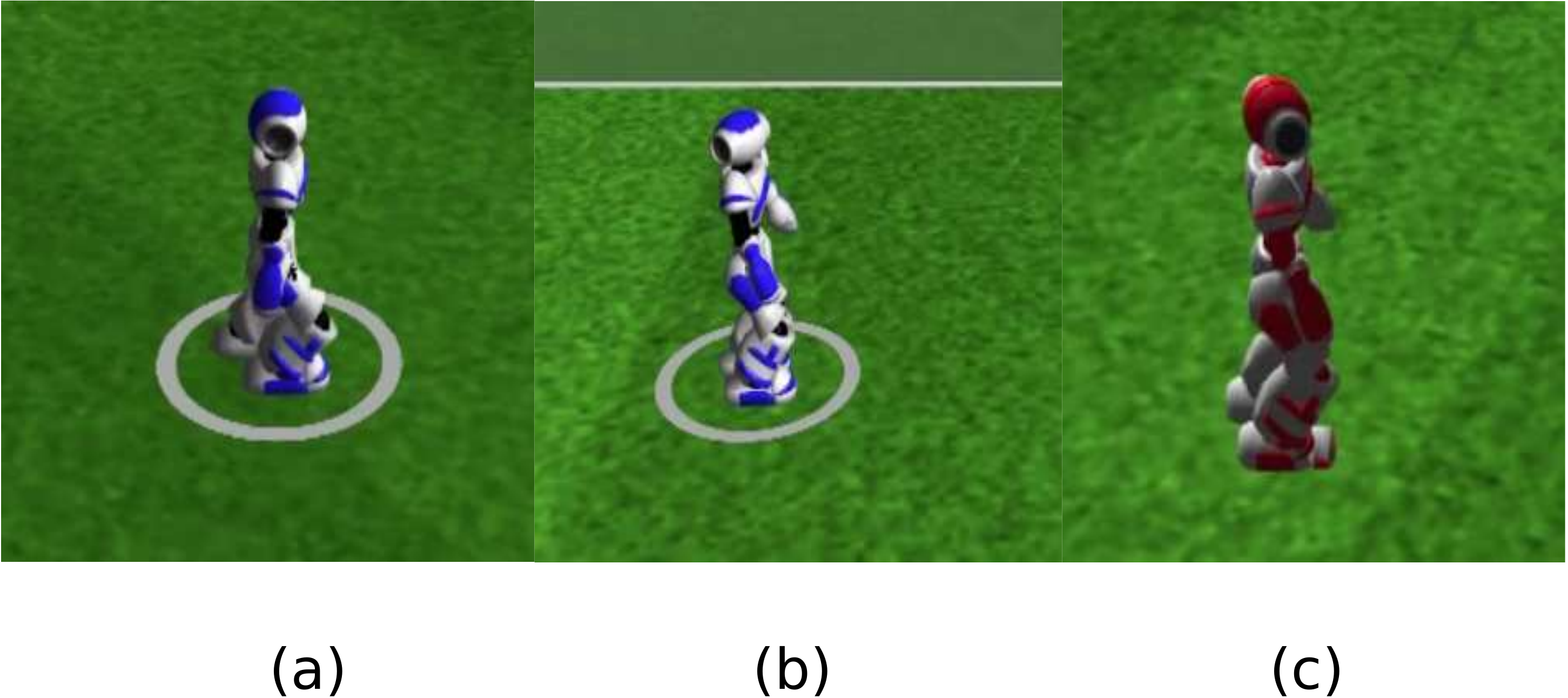}
\caption{Walking motions comparison. The figure (a) shows our agent in its regular walk. The figure (b) shows the same agent mimicking UT Austin Villa walk. The figure (c) is the UT Austin Villa agent itself, performing his own walking motion.}
\label{fig:walkings}
\end{figure}

Despite the facts above, the motion works well in a non-competitive scenario\footnote{\label{footnote_walk} Walk results video: https://streamable.com/m5w1d}, which is shown in the metrics collected in Forward Walk test scenario -- agent walking forward from the goal post until the center line of field -- in Table \ref{tab_walk} and the visual representation in Figure \ref{fig:walkings}.

\begin{table}[htbp]
\caption{Walk Comparison - Forward Walk}
\begin{center}
\begin{tabular}{|c|c|c|c|c|}
\hline
\textbf{Walk}&\multicolumn{4}{|c|}{\textbf{Statistics}} \\
\cline{2-5} 
\textbf{Type} &\multicolumn{2}{|c|}
{\textbf{Velocity \((m/s)\)}}
&\multicolumn{2}{|c|}{\textbf{Y Error \( (m) \) }} \\ 
\hline
 &
\textbf{\textit{Mean}} &
\textbf{\textit{Std}} & \textbf{\textit{Mean}} & \textbf{\textit{Std}} \\
\hline
Original Walk & 0.87 & 0.01 & - & -  \\
\hline
Learned Walk & 0.23 & 0.01 & 0.96 & 2.63 \\
\hline
\end{tabular}
\label{tab_walk}
\end{center}
\end{table}

\subsection{Other motions}

This same framework was used to learn other keyframe motions originated by our agent itself, such as the get up motion. As the cases previously described, the resultant neural network was capable of mimicking the keyframe including its interpolation. Hence, all of our keyframe motions can be replaced by neural motions with similar performance.

However, the huge improvement of this method is about mimicking other teams motions. In Soccer 3D environment, movements like kick and walking have giant impact in the team's performance. With this learning framework, our agent is able to mimic multiples movements from several teams.

As example, we collected data from UT Austin Villa kick, which was originally optimized using Deep Reinforcement Learning techniques \cite{mcalpine2017}. Our agent learned this kick without any additional optimization strategy: we just used samples collected from the modified server.

\section{Conclusion and Future Work}
\label{sec:conclusion}
In this work, we presented a method for learning humanoid robot movements using datasets composed of joint values at each time instant. The learning framework provided was capable to learn several types of motion, including walk and kick, without any change in network architecture or hyperparameters. Moreover, the learned motions have similar performance to the original ones. Furthermore, this framework was able to learn other teams motions, without any knowledge about the underlying implementation -- only using the joints values provided by a modified version of the server. This is a huge improvement in terms of getting improved motions, as our agent can mimic other teams motions using this machine learning technique.

As future work, we plan to apply Deep Reinforcement Learning algorithms to obtain faster and more robust kicks, using as ``seed'' the neural networks obtained in this work. Another track to be followed is to transfer the learning of this network to a new network that represents the motion policy itself (i.e a network which has as inputs the current state of the robot, including joint and link states, besides the current time instant), and optimize this motion policy in order to get a closed-loop walking and kick motion that can correct itself. As a long term goal, we intend to create model-free kick and walking engines.

\section{Acknowledgements}
We thank our sponsors ITAEx, Altium, Intel, Mathworks, Metinjo, Micropress, Poliedro, Polimold, Poupex, FHC, Rapid, and Solidworks. We are also grateful to ITA for supporting our work.

We would also like to show our gratitude to Patrick MacAlpine from UT Austin Villa team for sharing their ideas and code regarding the Soccer 3D simulation server modification.

Finally, we would like to acknowledge all the ITAndroids team, especially Soccer 3D simulation team members for the hard work in the development of the base code.

\bibliographystyle{unsrt}
\bibliography{references}

\end{document}